\documentclass[]{interact}

\usepackage[numbers,sort&compress]{natbib}
\renewcommand\bibfont{\fontsize{10}{12}\selectfont}

\usepackage{graphicx}
\usepackage{mathptmx}      
\usepackage{latexsym,amsmath,amssymb}
\usepackage{subfigure,multirow}
\usepackage{array}
\usepackage{booktabs}
\usepackage{caption}
\usepackage{makecell,booktabs,footnote}
\usepackage[table]{xcolor}
\usepackage{bm}
\usepackage[ruled,vlined]{algorithm2e}
\usepackage{threeparttable}
\usepackage{url}
\usepackage[misc]{ifsym}
\usepackage{color}
\definecolor{darkblue}{rgb}{0,0.1,0.7}

\usepackage{hyperref}
\hypersetup{colorlinks,
            linkcolor=darkblue,
            anchorcolor=darkblue,
            citecolor=darkblue}
\usepackage{rotfloat}

\newcommand{\tabincell}[2]{\begin{tabular}{@{}#1@{}}#2\end{tabular}}

\theoremstyle{plain}

\theoremstyle{definition}

\theoremstyle{remark}

\begin{document}


\title{Batch Sequential Adaptive Designs for Global Optimization}

\author{
\name{Ning Jianhui\textsuperscript{a} and Xiao Yao\textsuperscript{b}\thanks{CONTACT Xiao Yao. Email: xystatistics@mails.ccnu.edu.cn} and Xiong Zikang\textsuperscript{a}}
\affil{\textsuperscript{a}School of Mathematics and Statistics, Central China Normal University, Wuhan, China}  
\textsuperscript{b}School of Statistics and Mathematics, Zhongnan University of Economics and Law, Wuhan, China 
}

\maketitle
\begin{abstract}
Compared with the fixed-run designs, the sequential adaptive designs (SAD) are thought to be more efficient and effective. Efficient global optimization (EGO) is one of the most popular SAD methods for expensive black-box optimization problems. A well-recognized weakness of the original EGO in complex computer experiments is that it is serial, and hence the modern parallel computing techniques cannot be utilized to speed up the running of simulator experiments.  For those multiple points EGO methods, the heavy computation and points clustering  are the obstacles. In this work, a novel batch SAD method, named `` accelerated EGO ", is forwarded by using a refined sampling/importance resampling (SIR) method to search the  points with large expected improvement (EI) values. The computation burden  of the new method  is much lighter, and the  points clustering  is also avoided. The efficiency of the proposed SAD is validated by nine classic test functions with dimension from 2 to 12. The empirical results show that the proposed algorithm indeed can parallelize original EGO, and gain much improvement compared against the other parallel EGO algorithm especially under high-dimensional case. Additionally, we also apply the new method to the hyper-parameter tuning of Support Vector Machine (SVM). Accelerated EGO  obtains comparable cross validation accuracy with other methods and the CPU time can be reduced a lot due to the parallel computation and sampling method.
\end{abstract}

\begin{keywords}
Sequential Adaptive Designs; Global Optimization; Computer Experiment; Expected Improvement; Hyper-parameter Optimization
\end{keywords}

\section{Introduction}
\label{sec:intro}
As the advent of new technologies, the computer experiments have become very popular, more and more experiments are simulated in a mathematical model. Considering the complexity of the simulator model, some emulator/surrogate models, such as the Kriging model, have been proposed to approximate the real relationship between the input factors  and outputs.  Since the computation of computer experiments can be very  heavy, a good experimental design is important. As we  know the most popular designs for computer experiments are the space-filling designs, such as the Latin hypercube design (LHD) and Uniform design (UD). A major flaw of the fixed-run space-filling designs is that all of the experimental points are fixed before any experiments are launched, and make no use of  the information gained from the surrogate model.  The other class of designs, called sequential adaptive designs (SAD), are initialized from a small size of space-filling design, and then updated sequentially and design points are added based on the new information/features of the surrogate model. It has been believed that the SAD can be more efficient and effective than fixed-point designs \citep{LN2008, LMW2010, Quan2014}. Some great SAD methods have been developed for different objective oriented experiments, such as the global optimization, level surface estimation (LS), percentile estimation (PE), and global fit (GF). In this paper, we only consider the SAD for global optimization.

EGO is a widely used SAD method for expensive black-box global optimization problems, which starts from a space-filling design and then sequentially updates the design one point a time\citep{jones1998}. The new point is gained by maximizing the so called  `expected improvement (EI)' function. Under certain assumption the iterates from this method are dense, that is, this method does find the global optimum \citep{jones2001}. It can be shown that EI criterion integrates the information of predicted means and standard deviations based on surrogate model, i.e., EI represents a trade-off sampling between promising and uncertain zones \citep{Ginsbourger2010}. Due to its excellent performance on global optimization, the EI crieterion has been generalized for some more complex optimization occasions, for example the multi-fidelity optimization \citep{He2017}.

To efficiently take the advantages of parallel computation facilities to accelerate the running of computer experiments, some innovative researches have been proposed. Actually, while \cite{schonlau1997} proposed the single point EI,  multi-points EI ($q$-EI) was also suggested for batch-sequential optimization. However targeting multiple points at the same time involves numerous numerical computation that are extremely time-consuming.  \cite{Ginsbourger2010} subsequently suggested two heuristic strategies, \textit{Kriging Believer} and \textit{Constant Liar} to deal with this issue. These two strategies obtain $q$ updating points via sequentially optimizing EI $q$ times per cycle with an artificially updated Kriging model. Concretely,  Kriging Believer is to replace the response values by the Kriging mean predictor and Constant Liar is to take an arbitrary constant value $L$ set by the user (usually the minimum or maximum of all currently available observations). However, the computational cost for these algorithms is still ``expensive", and Kriging Believer may get trapped in a non-optimal region for many iterations \citep{Ginsbourger2010}. Then an explicit formula allowing a fast and accurate deterministic approximation of $q$-EI was given by \cite{Cl2013Fast}, for reasonably low values of $q$ (typically, less than 10). On the other hand, \cite{PWV2008, Quan2014} suggested some algorithms to reduce the computation burden and avoid the points clustering.

In summary, these multi-points EGO might be good batch SAD for global optimization, although the cost for generating the batch of points is time consuming and some methods may lead to points clustering, which is the motivation of our work. We proposed the accelerated EGO algorithm by embedding a refined resampling way  in order to obtain the batch of points, instead of employing the optimization techniques. The key advantage of this sampling method is that the computation burden can be relieved by a lot, and hence the speed of algorithm is accelerated. That's the reason of the name ``accelerated EGO". Meanwhile the resampling process is done with a randomized quasi Monte Carlo points set, which can avert the failure resulting from points clustering. 

The remainder of the paper is organized as follows: In \autoref{sec:EGO} the Kriging model and original EGO  are introduced briefly, and then the accelerated EGO  is described in detail in \autoref{sec:AEGO}; Some numerical comparisons among the new algorithm,  ordinary EGO, and parallel EGO are presented in \autoref{sec:simu}; In \autoref{sec:Hyper}, the application of the proposed algorithm to hyper-parameter optimization of support vector machine is discussed; finally the main conclusions are made in \autoref{sec:con}.

\section{EGO algorithm}\label{sec:EGO}
Let $D=[\bm a,\bm b]=[a_1,b_1]\times\cdots\times[a_s.b_s]$ be a rectangular experimental region in $R^s$ and let $f(\bm{x})$ be the simulator model whose explicit expression is unknown or known but rather complex on the region $D$.
Then there exists a black-box optimization problem for finding ${\bm{x}}^*$ and $M$ such that

\begin{equation}
M=f({\bm{x}}^*)=\min_{\bm x\in D}f(\bm{x}).\label{eq:problem}
\end{equation}
$M$ is the global minimum of $f(\bm{x})$ on $D$, and $\bm{x}^*$ is called a minimum point on $D$.

This black-box optimization can be formed as the term of computer experiments. Suppose $\bm{x}_i, i=1,\cdots,n$ are experimental points over domain $D$, and $y_i=f(\bm{x}_i)$ is the output corresponding to each $\bm{x}_i$. Then a surrogate model $\hat{f}(\bm{x})$ is fitted based on the input-output data and can be taken the place of objective function $f(\bm{x})$ to do the optimization.

\subsection{A brief introduction to Kriging}
The Kriging model was proposed by the South African geologist, D.G. Krige, in his master's thesis \cite{Krige1951} on analyzing mining data and systematically introduced to model computer experiments by \cite{sacks1989}. A comprehensive introduction can refer to \cite{Santner2018,Roustant2012,fang2018}.

The Kriging model can be expressed as

\begin{equation}
y(\bm{x})=\sum_{j=1}^L\beta_jB_j(\bm{x})+z(\bm{x})\label{eq:kriging-model},
\end{equation}
where $\{B_j(\bm{x}),j=1,\cdots,L\}$ is a chosen basis over the experimental domain, and $z(\bm{x})$ is a Gaussian process with mean zero, variance $\sigma^2$, and correlation function $r(\bm{x}_i,\bm{x}_j|\kappa)$, $\kappa$ denote the parameters in various correlation functions. 

The parameters $\bm{\beta}$, $\sigma^2$ and $\kappa$ in the Kriging model can be estimated by maximizing the likelihood function. Then under the normality assumption on $z(\bm{x})$, the best linear unbiased prediction at point $\bm{x}$ is given by

\begin{equation}\label{eq:kriging-mean}
\hat{y}(\bm{x})=\bm{b}(\bm{x})\hat{\bm{\beta}}+\bm{r} (\bm{x})R^{-1}(\bm{y}-\bm{B}\hat{\bm{\beta}}),
\end{equation}
and the uncertainty quantification of this predictor is

\begin{equation}\label{eq:kriging-sd}
s^2(\bm{x})=\hat{\sigma}^2\left[1-(\bm{b}'(\bm{x}),\bm{r}'(\bm{x}))\left(
\begin{matrix}
\bm{0} & \bm{B}'\\
\bm{B} & R
\end{matrix}\right)^{-1}\left(
\begin{matrix}
\bm{b}(\bm{x})\\
\bm{r}(\bm{x})
\end{matrix}\right) \right],
\end{equation}
where 
 $\bm{y}=(y_1,\cdots,y_n)'$, $R$ is an $n\times n$ matrix whose $(i,j)$-element is $r(\bm{x}_i,\bm{x}_j|\hat{\kappa})$, $\bm{b}(\bm{x})=(B_1(\bm{x}),\cdots,B_L(\bm{x}))'$, $r(\bm{x})=(r(\bm{x}_1,\bm{x}|\hat{\kappa}),\cdots,r(\bm{x}_n,\bm{x}|\hat{\kappa}))'$ and

$$\bm{B}=\left(
\begin{matrix}
B_1(\bm{x}_1) & \cdots & B_L(\bm{x}_1)\\
B_1(\bm{x}_2) & \cdots & B_L(\bm{x}_2)\\
\cdots & \cdots & \cdots \\
B_1(\bm{x}_n) & \cdots & B_L(\bm{x}_n)
\end{matrix} \right).$$

\subsection{EGO algorithm}
The original EGO algorithm is an efficient single point SAD method for finding global optimum. In each design stage, it involves to computing the EI function for $\forall\bm{x}\in D$, whose analytical expression is as follows:

\begin{equation}
EI(\bm{x})=\left(\min_{i=1,\cdots,n}y_i-\hat{y}(\bm{x})\right)\Phi\left(\frac{\min_{i=1,\cdots,n} y_i-\hat{y}(\bm{x})}{s(\bm{x})}\right)+s(\bm{x})\phi\left(\frac{\min_{i=1,\cdots,n} y_i-\hat{y}(\bm{x})}{s(\bm{x})}\right)\label{eq:EI-analytical}
\end{equation}
where $\phi$ and $\Phi$ are the density and distribution of the standard normal distribution respectively, $\hat{y}(\bm{x})$ and $s^2(\bm{x})$ are the prediction \eqref{eq:kriging-mean} and predicted variance \eqref{eq:kriging-sd} derived from Kriging model, $y_i,\ i=1,\cdots,n$ are the output of experimental points.

Then the point $\bm{x}^{new}$ with the largetest EI value in domain $D$, i.e.
\begin{equation}
\bm{x}^{new}=\arg\max_{\bm{x}\in D}EI(\bm{x})\label{eq:x_new}
\end{equation}
is selected to be the next stage experimental point.

The procedure of this  sequential experiment  for global optimization \eqref{eq:problem} can be formed as Algorithm 1.

\begin{algorithm}[h]
	\caption{EGO Algorithm}
	\LinesNumbered 
	\KwIn{an initial design $\mathcal{P}=\{\bm{x}_i, i=1,\cdots,n\}$}
	\KwOut{the best approximation $\bm{x}^{best}$ and $M^{best}$}
    Evaluate $y_i=f(\bm{x}_i)$, set $\bm{y}=(y_1,\cdots,y_n)$, and fit the Kriging model\; 
	\While{the stop condition is not met}{
     Find a new point $\bm{x}^{new}$ in the experimental domain $D$ maximizing the $EI(\bm{x})$, i.e.,
     	$$\bm{x}^{new}=\arg\max_{\bm{x}\in D}EI(\bm{x});$$\\
     Evaluate $f(\bm{x}^{new})$, and set $\mathcal{P}:=\mathcal{P}\cup \{\bm{x}^{new}\}$, $\bm{y}:=\bm{y}\cup \{f(\bm{x}^{new})\}$\;
     Re-fit the Kriging model with the design points $\mathcal{P}$.
	}
\end{algorithm}

\section{Accelerated EGO}\label{sec:AEGO}
\subsection{The randomized quasi-random SIR process}
As discussed in \autoref{sec:intro}, the big obstacles of multi-points or parallel EGO are the heavy computation and points clustering. Here, we propose a computationally easier algorithm to parallelize EGO algorithm.

Actually, there are two key properties of the updating points in parallel EGO algorithms. Firstly, the chosen points should be the peak of EI function, or at least with large EI values. Secondly, these points should be distributed as uniformly as possible around the target area. We employ the randomized quasi-random sampling/importance resampling (RQSIR) process to target the updating points and ensure the two properties as well. In the process of RQSIR, the EI function is seen as the kernel function of a probability density function, i.e.
$$\bm{x}\sim g(\bm{x})\varpropto EI(\bm{x}),$$
and updating points are generated from $g(\bm{x})$ by RQSIR method. The process of RQSIR includes  the following three steps:

\begin{enumerate}
	\item \textbf{Choosing a candidate points pool} $\mathcal{Q}=\{\bm{z}_i=(z_{i1},\cdots,z_{is}), i=1,\cdots,m\}$.

It is well known that quasi-Monte Carlo (QMC) method is a efficient method for high-dimensional integration approximation. This method uses a set of points, called QMC sequence, to approximate the integration \citep{Niederreiter1992}. The QMC points are also called low discrepancy sequence, they are uniformly distributed in the integrated  domain. There are many ways to genrerate them, such as the Halton and Sobol sequence, which can be directly generated by R package ``randombooltox" \citep{rand2019}; Or the uniform design, which can be obtained from R package ``UniDOE" \citep{unidoe2018}. Of course, other space-filling designs, like \cite{Mak2017},  also can be employed.
	\item \textbf{Randomizing the candidate points pool} $\mathcal{Q}$ to get $\mathcal{Q}_R$:

Unlike the Monte Carlo points, the QMC points are certainly deterministic. If adopting the points in this QMC points  pool directly in the third ``resampling" step,  only these points have the chance be chosen. Hence we employ the Randomized QMC in this step to introduce the randomness in, and keep their distribution construction invariant as well. There are several randomization methods \cite{Ecuyer2002}. Here, we only employ the simplest way, called random shift.

\begin{itemize}
 \item Firstly, an $s$-dimensional random vector $\Delta\sim U(D)$ is generated, and then shift the whole points set $\mathcal{Q}$ to get $$\mathcal{Q}^*=\{\bm{z}^*_i=\bm{z}_i+\Delta;\; \bm{z}_i\in \mathcal{Q},i=1,\cdots,m\},$$
     where $D=[\bm a,\bm b]=[a_1,b_1]\times\cdots\times[a_s.b_s]$ is the experimental domain.
 \item Secondly, an analogized modular operation is put on $\mathcal{Q}^*$ to get $\mathcal{Q}_R=\{\tilde{\bm{z}}_i=(\tilde{z}_{i1},\cdots, \tilde{z}_{is})$, $i=1,\cdots,m\}$, i.e.

\begin{equation}
\tilde{z}_{ik} = \begin{cases}
a_k+(z_{ik}^*-b_k), & \text{if } z_{ik}^*>b_k,\\
z_{ik}^*, & \text{if } a_k\leq z_{ik}^*\leq b_k,\\
b_k-(a_k-z_{ik}^*), & \text{if } z_{ik}^*<a_k
\end{cases}
\end{equation}
for $i=1,\cdots,m; \; k=1,\cdots,s$.
\end{itemize}
The randomize QMC process for 2-dimension domain $D$ can be visualized as shown in \autoref{fig:randomization}.

\begin{figure}[h]
	\centering
	\subfigure[$\mathcal{Q}$]{
		\includegraphics[width=3.8cm, height=4cm]{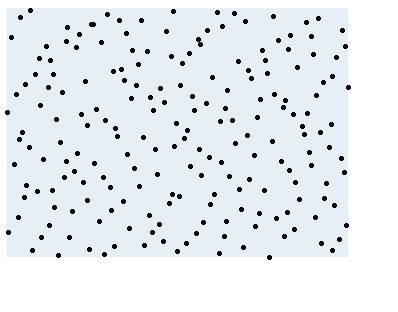}}
	\subfigure[$\mathcal{Q}^*$]{
		\includegraphics[width=3.8cm, height=4cm]{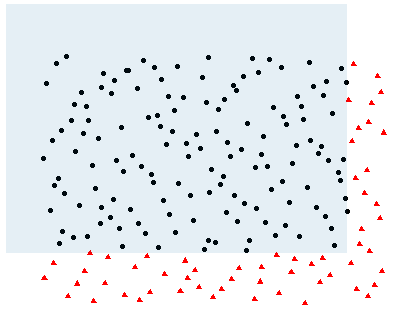}}
	\subfigure[$\mathcal{Q}_R$]{
		\includegraphics[width=3.8cm, height=4cm]{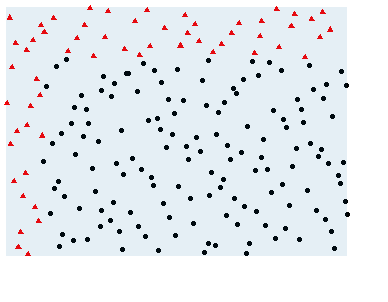}}
	\caption{The process of randomized quasi-Monte Carlo.[A Sobol sequence with 200 points, the shaded region is the experimental domain.]}
	\label{fig:randomization} 
\end{figure}

 \item \textbf{Resampling from the randomized candidate points pool} $\mathcal{Q}_R$ .

 At each iteration, the EI values of each $\tilde{\bm{z}}_i \in \mathcal{Q}_R, i=1,\cdots,m$ can be calculated according to \eqref{eq:EI-analytical}. Then the augmented new experimental points $\mathcal{P}^*=\{\tilde{\bm{x}}_1,\cdots,\tilde{\bm{x}}_J\} $ $(J\leqslant m)$ are sampled from the pool $\mathcal{Q}_R$ according to the weights
\begin{equation}
\varpi_i=\frac{EI(\tilde{\bm{z}}_i)}{\sum_{i=1}^mEI(\tilde{\bm{z}}_i)},\; i=1,\cdots,m.
\end{equation}

\end{enumerate}

\subsection{Accelerated EGO algorithm}
All in all, the RQSIR process helps to keep the diversity of candidates, ensures that every point in the experimental domain has chance to be chosen and can bring more exploration in the neighborhood of global optimum. In addition, these updating points can keep certain distance with the existing points. For the global optimization problem \eqref{eq:problem}, by nesting the RQSIR process described above, the new batch sequential adaptive design method, accelerated EGO, can be stated as in Algorithm 2.

\begin{algorithm}[h]
	\caption{Accelerated EGO Algorithm}
	\LinesNumbered 
	\KwIn{an initial design $\mathcal{P}=\{\bm{x}_i, i=1,\cdots,n\}$ and a candidate points pool $\mathcal{Q}=\{\bm{z}_i, i=1,\cdots,m\}$}
	\KwOut{the best approximation $\bm{x}^{best}$ and $M^{best}$ }
	Evaluate $y_i=f(\bm{x}_i)$, set $\bm{y}=(y_1,\cdots,y_n)$, and fit the Kriging model\; 
	\While{the stop condition is not met}{
		Find a new point $\bm{x}^{new}$ in the experimental domain $D$ maximizing the $EI(\bm{x})$, i.e.,
		$$\bm{x}^{new}=\arg\max_{\bm{x}\in D}EI(\bm{x});$$ \\
		Randomization: randomize $\mathcal{Q}=\{\bm{z}_1,\cdots,\bm{z}_m\}$ to get $\mathcal{Q}_R=\{\tilde{\bm{z}}_1,\cdots,\tilde{\bm{z}}_m\}$\;
		Resampling: draw $J$ points $\mathcal{P}^*=\{\tilde{\bm{x}}_1,\cdots,\tilde{\bm{x}}_J\} (J\leqslant m)$ from $\mathcal{Q}_R$ according to the weights
		$$\varpi_i=\frac{EI(\tilde{\bm{z}}_i)}{\sum_{i=1}^{m}EI(\tilde{\bm{z}}_i)}, i=1,\cdots,m$$
		where $EI(\tilde{\bm{z}}_i), i=1,\cdots,m$ is calculated based on current Kriging model\;
		Evaluate $f(\bm{x}^{new})$ and $\{f(\mathcal{P}^*)\}$, and set $\mathcal{P}:=\mathcal{P}\cup \{\bm{x}^{new}\}\cup \mathcal{P}^*$, $\bm{y}:=\bm{y}\cup \{f(\bm{x}^{new})\}\cup \{f(\mathcal{P}^*)\}$\;
		Re-fit the Kriging model with the design points $\mathcal{P}$.
	}
\end{algorithm}

\subsection{Toy example}
Here, the 2-dimensional Ackley function is taken as a demonstrative example to show the procedure of the accelerated EGO  and do the sensitive analysis about the size of initial QMC points pool.
The Ackley function
\begin{equation}
f(x_1,x_2)=-20\exp\left(-\frac{1}{5}\sqrt{\frac{1}{2}(x_1^2+x_2^2)}\right)-\exp\left(\frac{1}{2}(\cos(2\pi x_1)+\cos(2\pi x_2))\right)+20+e
\end{equation}
with $-2\leq x_1,x_2\leq2$, has several local optima around the unique global minimum $\bm{x}^*=(0,0), f(\bm{x}^*)=0$.

\begin{figure*}[t]
	\centering
	\subfigure[initial]{
		\includegraphics[width=4.5cm, height=4cm]{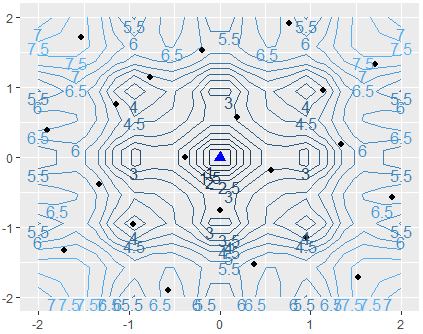}}
	\subfigure[iteration1]{
		\includegraphics[width=4.5cm, height=4cm]{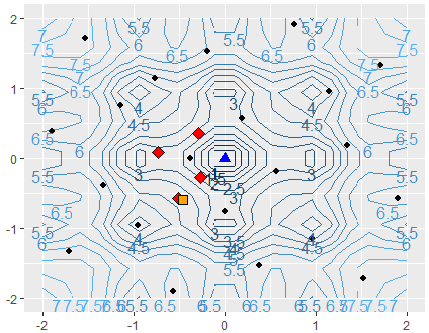}}
	\subfigure[iteration2]{
		\includegraphics[width=4.5cm, height=4cm]{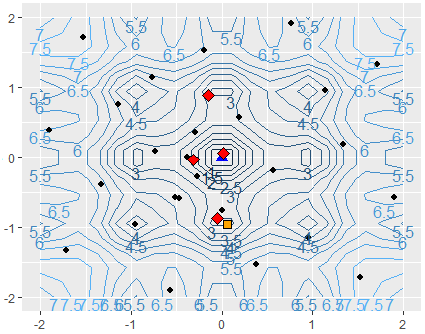}}
    \subfigure[iteration3]{
		\includegraphics[width=4.5cm, height=4cm]{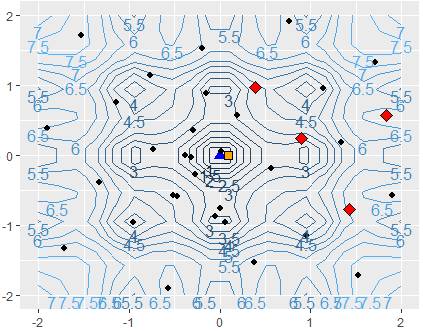}}
    \subfigure[iteration4]{
		\includegraphics[width=4.5cm, height=4cm]{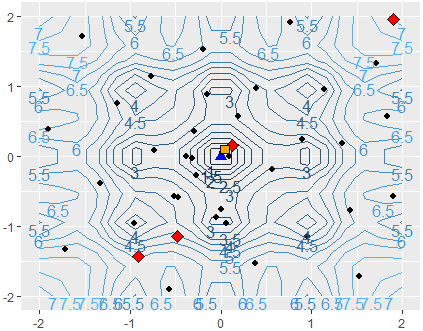}}
    \subfigure[iteration5]{
		\includegraphics[width=4.5cm, height=4cm]{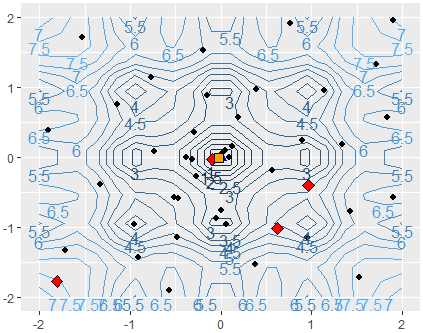}}
	\caption{The search process of Ackley function. [The black dots ``$\cdot$" are the experimental points; the blue triangle ``\textcolor{blue}{$\blacktriangle$}" is the global minimum point; the green rectangle ``\textcolor{orange}{$\blacksquare$}" is the largest EI point and the red lozenge ``\textcolor{red}{$\blacklozenge$}" are the remaining selected points at each iterations]}
	\label{fig:ackley2} 
\end{figure*}

Firstly, the QMC points pool for RQSIR is set to be a Sobol sequence with $100$ candidates and $5$ points are required to update per stage, i.e, the largest EI point and $4$ points sampling from the ramdomized QMC points set. The accelerated EGO needs $5$ cycles to find the global optimum with the stop condition being $\left|M^{best}-M\right|<10^{-2}$. The search process can be seen from \autoref{fig:ackley2}. At the initial stage, a uniform design with $21$ points is scattered over the experimental domain. At stage $1$-$5$, the point with the largest EI value is obtained and $4$ points with relative large EI values are sampled from the $100$ candidates. It can be seen that the sampling points are not flocked together and  bring more exploration to the subdomain global optimum located at.

Then, the sensitive analysis about the size of QMC points pool can be done based on this Ackley function. We choose the Sobol sequence with size $m=20, 50, 100, 150$ respectively to be the candidates. 3 and 5 updating points are selected respectively per stage by the accelerated EGO algorithm and the stop condition is set to be $\left|M^{best}-M\right|<10^{-2}$. In order to average out the bias caused by some latent factors, $100$ repetitions are done for each case. \autoref{tab:pool_size} lists the number of cycles required to satisfy the stop condition for each combination of pool size $m$ and the number of updating points, where it can be observed that the number of cycles are not too sensitive about the size of QMC points pool. Of course the size $m$ can not be too small and is relative to the dimension $s$ of the optimization problem. Generally speaking, we recommend that the size $m$ should set to be $50s\sim100s$ or so.

\begin{table}[htbp]
	\centering
	\caption{The influence of different size of QMC points pool}
    \linespread{1.5}
    \label{tab:pool_size}
\begin{threeparttable}
	\begin{tabular}{ccccccc}
	  \hline
      \multirow{2}*{\tabincell{c}{pool size \\ $m$}} & \multicolumn{3}{c}{5-point} & \multicolumn{3}{c}{10-point} \\ \cmidrule(r){2-4} \cmidrule(r){5-7}
       &median & mean & s.t.d & median & mean & s.t.d  \\ \hline
      20 & 5 & 6.65 & 3.41 & 5 & 5.79 & 2.10 \\
      50 & 5 & 6.22 & 3.17 & 4 & 4.69 & 1.25 \\
     100 & 4 & 4.73 & 4.03 & 4 & 5.12 & 2.37 \\
     150 & 5 & 5.92 & 3.33 & 4 & 5.06 & 2.35 \\
     \hline
	\end{tabular}
    \begin{tablenotes}
    \footnotesize
    \item[1] The ``$q$-point" is accelerated EGO with $q$ updating points per cycle.
    \end{tablenotes}
\end{threeparttable}
\end{table}

\section{Numerical comparisons} \label{sec:simu}

In this section, we use nine classic multi-modal test functions with different dimensions to evaluate the performance of the proposed algorithm, accelerated EGO. The majority of these test functions are employed by other refined EGO \citep{huang2006,mehdad2018,he2018} and parallel EGO \citep{Ginsbourger2010,Viana2013} to validate algorithms' efficiency. As the bench marker, the performance of original EGO and the most popularly used parallel EGO---Constant Liar algorithm are also included. 

\subsection{Test functions and algorithm setting}
The nine test functions are described as follows:

1. \textbf{Branin function} ($s=2$) 
\begin{equation*}
f(x,y) = (y-\frac{5}{4\pi^2}x^2+\frac{5}{\pi}x-6)^2+10(1-\frac{1}{8\pi})\cos x+10
\end{equation*}
with $x\in [-5, 10], \; y\in [0,15]$. The global minimum $M=0.397887$ are located at $\bm{x}^*=(-\pi$, $12.275)$, $(\pi, 2.275)$ and $(9.42478, 2.475)$.

2. \textbf{Six-hump camel function} (SixCamel) ($s=2$) 
\begin{equation*}
f(x,y)= 4x^2-2.1x^4+x^6/3+xy -4y^2+4y^4
\end{equation*}
with  $-2\leq x\leq 2$, $-1\leq y\leq 1$. It has six extreme points, among them the global minimum are $\bm{x}^*=(0.0898,-0.7126)$, $(-0.0898,0.7126)$ and $M=-1.0316$.

3. \textbf{Modified Goldstein-Price function} (GoldPrice) ($s=2$) 
\begin{equation*}
\begin{split}
f(x,y)=&\frac{1}{2.427}[\log( [1+(x+y+1)^2(19-14x+3x^2-14y+6xy+3y^2)]\\
&[30+(2x-3y)^2(18-32x+12x^2+48y-36xy+27y^2)])-8.693]
\end{split}
\end{equation*}
with $-2\leq x,y\leq2$. The global minimum $M=-3.129126$ is at point $\bm{x}^*=(0,-1)$ with several local minima around it.

4. \textbf{SIN2 function} ($s=2$)
\begin{equation*}
f(x,y)= 1+\sin^2(x)+\sin^2(y)-0.1\exp(-x^2-y^2)
\end{equation*}
with $-5\leq x\leq 5$, $-5\leq y\leq 5$. It has many extreme points and the global minimum is $\bm{x}^*=(0, 0)$, $M=0.9$.

5. \textbf{Hartmann function} with $s=3$ (Hartmann3) and $s=6$ (Hartmann6) 
\begin{equation*}
f(\bm{x})=-\sum\limits_{i=1}^4\alpha_i\exp\left(-\sum\limits_{j=1}^sA_{ij}(x_j-P_{ij})^2 \right)
\end{equation*}
where $0\leq x_i\leq1, i=1,2,\cdots,s$ and $\alpha=(1.0, 1.2, 3.0, 3.2)^T$. The elements of matrices $\bm{A}$, $\bm{P}$ and the global minimum are shown in \autoref{tab:Hart}.

\begin{table}[h]
	\centering
	\caption{Global minimum and settings of Hartmann function}
    \linespread{3.5}
	\begin{tabular}{c|c}
		\hline
		Functions & Parameters \\
		\hline
		\tabincell{c}{Hartmann3\\$\bm{x}^*=(0.1146,0.5556, 0.8525) $\\$ M=-3.86278$}
		&$\bm{A} = \left(
		\begin{matrix}
		3.0 & 10 & 30 \\
		0.1 & 10 & 35 \\
		3.0 & 10 & 30 \\
		0.1 & 10 & 35
		\end{matrix} \right);$
		$ \bm{P} = 10^{-4}\left(
		\begin{matrix}
		3689 & 1170 & 2673 \\
		4699 & 4387 & 7470 \\
		1091 & 8732 & 5547 \\
		381 & 5743 & 8828
		\end{matrix} \right) $\\
		\hline
		\tabincell{c}{Hartmann6\\$\bm{x}^*=(0.2017,  0.1500, 0.4769, $\\$ 0.2753, 0.3117, 0.6573)$\\$M=-3.32237$}
		& \tabincell{c}{ $\bm{A} = \left(
			\begin{matrix}
			10 &   3 &   17 & 3.5 & 1.7 &  8 \\
			0.05 &  10 &   17 & 0.1 &   8 & 14 \\
			3 & 3.5 &  1.7 &  10 &  17 &  8 \\
			17 &   8 & 0.05 &  10 & 0.1 & 14
			\end{matrix} \right)$ \\
			$\bm{P} = 10^{-4}\left(
			\begin{matrix}
			1312 & 1696 & 5569 &  124 & 8283 & 5886 \\
			2329 & 4135 & 8307 & 3736 & 1004 & 9991 \\
			2348 & 1451 & 3522 & 2883 & 3047 & 6650 \\
			4047 & 8828 & 8732 & 5743 & 1091 &  381
			\end{matrix} \right) $ }\\
		\hline
	\end{tabular}
	\label{tab:Hart}
\end{table}

6. \textbf{Ackley function} with $s=10$ (Ackley10) 
\begin{equation*}
f(\bm{x})=-a\exp\left(-b\sqrt{\frac{1}{s}\sum_{i=1}^sx_i^2}\right)-\exp\left(\frac{1}{s}\sum_{i=1}^s\cos(cx_i)\right)+a+e
\end{equation*}
where $a =20, \; b=0.2, \; c=2\pi$, $-5.12\leq x_i\leq5.12, \; i =1,\cdots,s$. And the global minimum is $M=0$ at $\bm{x}^*=(0,\cdots,0)$.

7. \textbf{Levy function} with $s=10$ (Levy10) 
\begin{equation*}
f(\bm{x})=\sin^2(\pi w_1)+\sum_{i=1}^{s-1}(w_i-1)^2[1+10\sin^2(\pi w_i+1)]+(w_s-1)^2[1+\sin^2(2\pi w_s)]
\end{equation*}
where $w_i=1+\frac{x_i-1}{4},\; i=1,\cdots,s.$
And $-10\leq x_i\leq10$, the global optimum is $M=0$ at $\bm{x}^*=(0,\cdots,0)$.

8. \textbf{Trid function} with $s=12$ (Trid12) 
\begin{equation*}
f(\bm{x})=\sum_{i=1}^s(x_i-1)^2-\sum_{i=2}^sx_ix_{i-1}
\end{equation*}
where $-s^2\leq x_i\leq s^2$, $i=1,\cdots,s$. And the global optimum is $M=-s(s+4)(s-1)/6$ at $x_i=i(d+1-i)$, $i=1,\cdots,s$.

As for algorithm setting, the number of initial experimental points (uniform design) for the three algorithms and the size of QMC points pool (Sobol sequence) in each test functions are list in \autoref{tab:setting}. As \cite{Ginsbourger2010} shows, Constant Liar algorithm with the constant setting being the minimum of the response values can get the best results. Therefore, in this paper, Constant Liar algorithm with this setting is included, denoted by ``CL(min)" as \cite{Ginsbourger2010} does. And all the experiments run on a PC consisting of 3.60GHz Intel(R) Core(TM) i7-7700 processor and 16-GB memory. 

\begin{table}[h]
	\centering
	\caption{The number of points in initial design and Sobol pool}
	\begin{tabular}{ccc|ccc}
		\hline
		\tabincell{c}{Test \\ Function} & \tabincell{c}{initial \\design} & \tabincell{c}{Sobol\\ pool} & \tabincell{c}{Test\\ Function} & \tabincell{c}{initial \\design} & \tabincell{c}{Sobol\\ pool}  \\
		\hline
		Branin & 21 & 100 & Ackley10 & 100 & 750 \\
	    SixCamel & 21 & 100 & Levy10 & 100 & 750 \\
     GoldPrice & 21 & 100 & Trid12 & 120 & 1000\\
          SIN2 & 21 & 100 & \\
	 Hartmann3 & 35 & 150 & \\
     Hartmann6 & 65 & 300 & \\
	\hline
	\end{tabular}
	\label{tab:setting}
\end{table}

\subsection{Low-dimensional case}

For the test functions with $s=2$ to $6$, the comparing metric is the number of stages which the sequential design require to satisfy the stop condition $|M^{best}-M|<\epsilon$. The number of stages can measure the ability to parallelize original EGO. In addition, the number of evaluations is equal to $q$ (the number of updating points per stages) times the number of stages, so only the number of stages is counted. In order to eliminate the influence of other inessential factors, all the experiments are repeat 100 times under this low-dimensional case. The results (the mean, standard deviation and median of the number of stages) are listed in \autoref{tab:low_cycles} and \autoref{tab:low_time} shows the CPU time three algorithms need to find the optimum in the form of ``mean (standard deviation)" .

The results listed in \autoref{tab:low_cycles} indicate that the resampling method is efficient to generate  updating points on these test functions no matter how many updating points are needed per stage. For example, EGO algorithm needs  $13.89$ stages averagely to satisfy the error for Branin function, while the $12$-point accelerated EGO needs only $2.45$ iterations averagely. For SIN2 function, the comparison is $29.85$-$4.01$, and is $12.7$-$4.48$ for Hartmann6 function. As the number of updating points increases, the mean, standard deviation and median reduce therewith. For example, when the number of updating points increases from 4 to 12, the mean of the number of stages is from $5.94$ to $5.18$, the standard deviation is $1.94$-$1.48$ and the median is $5.5$-$4$ for Hartmann3 function. These show that the accelerated EGO  becomes more efficient and robust when more updating points are selected per stage.

\begin{sidewaystable*}
	\centering
	\caption{Optimization comparisons under low-dimensional case}
\linespread{5}
    \label{tab:low_cycles}
\begin{threeparttable}
	\begin{tabular}{cccccccccc}
		\hline
		\multirow{2}*{Test Function} & \multirow{2}*{ $\epsilon$} & \multirow{2}*{\tabincell{c}{Number of \\ iterations}} & \multirow{2}*{EGO} & \multicolumn{3}{c}{CL(min)} & \multicolumn{3}{c}{Accelerated EGO}\\
	\cmidrule(r){5-7}  \cmidrule(r){8-10}	
      & & &  & 4-point & 8-point & 12-point & 4-point & 8-point & 12-point\\
      \hline
      \multirow{3}*{Branin} & \multirow{3}*{$10^{-2}$} &  mean & 13.89 & 3.95 & 3.46 & 3 &4.04 & 2.89 & 2.45 \\
      & & s.t.d. & 1.48 & 1.17 & 0.50 & 0.00 &  1.59 & 0.76 & 0.50 \\
      & & median & 13 & 3 & 3 & 3 & 4 & 3 & 2\\ \hline
      \multirow{3}*{SixCamel} & \multirow{3}*{$10^{-3}$} &  mean & 9.58 & 3.60 & 2.71 & 2.60 & 4.61 & 3.50 & 2.78 \\
      & & s.t.d. & 1.13 & 0.83 & 0.55 & 0.49 & 1.43 & 0.85 & 0.59 \\
      & & median & 10 & 4 & 3 & 3 & 4 & 3 & 3 \\ \hline
      \multirow{3}*{GoldPrice} & \multirow{3}*{$10^{-2}$} &  mean & 56.70 & 21.70 & 18.98 & 15.78 & 20.32 & 17.84 & 13.84 \\
      & & s.t.d. & 31.27 & 11.48 & 10.65 & 10.84 & 13.17 & 10.73 & 8.55\\
      & & median & 57 & 22 & 18.5 & 13.5 & 18.5 & 15 & 11 \\ \hline
      \multirow{3}*{SIN2} & \multirow{3}*{$10^{-2}$} & mean & 29.85 & 8.45 & 5.00 & 3.96 & 8.68 & 5.33 & 4.01 \\
      & & s.t.d. & 6.59 & 1.51 & 0.73 & 0.62 & 2.22 & 1.04 & 0.88 \\
      & & median & 30.5 & 8.5 & 5 & 4 & 9 & 5 & 4 \\ \hline
      \multirow{3}*{Hartmann3} & \multirow{3}*{$10^{-4}$} & mean & 14.34 & 5.24 & 5.00 & 4.20 & 5.94 & 5.78 & 5.18 \\
      & & s.t.d. & 3.42 & 0.65 & 3.73 & 1.95 & 1.94 & 1.66 & 1.18 \\
      & & median & 13 & 5 & 4 & 4 & 5.5 & 5 & 4\\ \hline
      \multirow{3}*{Hartmann6} & \multirow{3}*{$10^{-1}$} & mean & 21.7 & 6.04 & 4.83 & 4.06 & 5.62 & 4.91 & 4.48 \\
      & & s.t.d. & 1.83 & 0.47 & 0.38 & 0.34 & 2.27 & 2.77 & 2.12 \\
      & & median & 15 & 6 & 5 & 4 & 5 & 4 & 3 \\
	\hline
	\end{tabular}
    \begin{tablenotes}
    \linespread{1}
        \footnotesize
        \item[1] The ``$q$-point" means the number of updating points per iteration for CL(min) and accelerated EGO.
      \end{tablenotes}
\end{threeparttable}
\end{sidewaystable*}

However, the improvement of accelerated EGO is not strictly linear. That is to say, the reduction of iterations is not strictly in proportion to the number of updating points at each stage. For example, 4-point accelerated EGO algorithm speeds up $3.44$ ($\approx29.85/8.68$) times for SIN2 function, which should be 4 times for a linear speedup algorithm. Since the speedup of accelerated EGO algorithm is below linear, the number of evaluations that this algorithm requires to reach the error would be greater than that of ordinary EGO. But with the ability of parallel computation, this could not be a serious problem. 

The comparisons between accelerated EGO and CL(min) algorithms under low-dimensional case are also shown in \autoref{tab:low_cycles}. It can be observed that accelerated EGO performs similarly to CL(min) algorithm from the number of stages aspect for most test functions. For example, the number of stages of accelerated EGO is $0.18$ more than CL(min) averagely and the median of stages is the same to be $3$ for SixCamel function with 12 updating points. And for Hartmann6 function, when 8 points are selected at each stage, the mean of stages is $4.83$-$4.91$ for CL(min) and accelerated EGO algorithms. However, it seems that accelerated EGO has more advantages than CL(min) algorithm on test functions with complex surface. For example, GoldPrice function is thought to be hard to optimize with a sharp region of global minimum. The accelerated EGO performs much better on this test function no matter how many updating points are selected. The comparisons of average number of stages are $21.70$-$20.32$, $18.98$-$17.84$ and $15.78$-$13.84$ for CL(min) and accelerated EGO algorithms with 4, 8 and 12 updating points per stage.

\begin{table}[htbp]
	\centering
	\caption{CPU time under low-dimensional case}
    \label{tab:low_time}
\begin{threeparttable}
	\begin{tabular}{ccccccc}
		\hline
	  \multirow{2}*{\tabincell{c}{Test\\functions}} & \multicolumn{3}{c}{CL(min)} & \multicolumn{3}{c}{Accelerated EGO}\\
	\cmidrule(r){2-4}  \cmidrule(r){5-7}	
      & 4-point & 8-point & 12-point & 4-point & 8-point & 12-point\\
      \hline
      Branin & 5.9 (2.6) & 10.2 (2.5) & 12.3 (0.5) & 2.3(1.1) & 1.6 (0.6) & 1.2 (0.4) \\
      SixCamel & 9.5 (2.2) & 12.2 (2.8) & 19.0 (4.4) & 3.6(1.3) & 2.8 (0.8) & 2.3 (0.6) \\
      GoldPrice & 44.2 (26) & 95.1( 61) & 103.8 (92) & 14.9 (11) & 19.1 (14) & 15.8 (16)\\
      SIN2 & 26.7 (5.8) & 30.12 (5.7) & 33.8 (7.1) & 8.0 (2.4) & 5.2 (1.6) & 3.9 (1,1)\\
      Hartmann3 & 20.2 (4.1) & 35.0 (29.7) & 42.3 (21.8) & 10.1 ( 4.1) & 9.5 (2.9) & 8.1 (2.4)\\
      Hartmann6 & 68.7 (12) & 98.9 (17) & 106.7 (15) & 33.8 (23) & 34.0 (32) & 38.5 (57)\\
	\hline
	\end{tabular}
    \begin{tablenotes}
    \linespread{1}
        \footnotesize
        \item[1] The CPU time is in the form of ``mean (standard deviation)".
    \end{tablenotes}
\end{threeparttable}
\end{table}

Another significant superiority, the CPU time needed to reach the given error of test functions, can be obtained for accelerated EGO comparing with CL(min) algorithm. When the ability of parallelizing EGO algorithm (the number of stages) is similar for CL(min) and accelerated EGO algorithm, the most noticeable issue is that the complexity of targeting the multiple updating points. As \autoref{tab:low_time} shows, accelerated EGO is a faster method to optimize all the test functions no matter how many updating points are used at each stage. For example, the CPU time is $3.6$ seconds for 4-point accelerated EGO to satisfy the stop condition of SixCamel function, while the CL(min) algorithm runs $9.5$ seconds averagely. This advantage can be more significant when more updating points are selected per stage. For example,  when 4 updating points are added per stage for SIN2 function, the CPU time of CL(min) is averagely $3.3$ ($\approx 26.7/8.0$) times as much as accelerated EGO, while for 12-point situation, the ratio is $8.6$ ($\approx33.8/3.9$). This phenomenon indicates that with accelerated EGO algorithm, the computational burden for targeting multiple points can be eased a lot.

\subsection{High-dimensional case}
In this subsection, the ``Ackley10", ``Levy10" and ``Trid12" test functions are studied. Under the high-dimensional case, it is hard to reach the given error for the test function with limited evaluations. Thus, the stop condition is set to be the number of exerimental runs in total. And the comparing metric is the minimum value the three algorithms can obtain with the given number of evaluations.

For ``Ackley10" and ``Levy10" test functions, the total runs is set to be $250$. Subtracting the runs of initial design (a uniform design with $100$ points), $150$ updating points are available. And for ``Trid" test function, the number of updating points is $150$, and with a $120$- point initial design the total runs are $270$. EGO algorithm sequentially run these $150$ updating points one by one for these three test functions. Parallel algorithms, CL(min) $\&$ accelerated EGO,  run $10$ and $15$ updating points respectively per stage. All the simulations in this case are repeated $10$ times. The results (the minimum value the algorithms can get and the CPU time) are recorded in \autoref{tab:high_result} in the form of ``mean (deviation)". 

\begin{table}[h]
	\centering
	\caption{Optimization comparisons under high-dimensional case}
    \label{tab:high_result}
\begin{threeparttable}
	\begin{tabular}{ccccccc}
	\hline
	\multirow{2}*{\tabincell{c}{Test\\functions}} & &\multirow{2}*{EGO} & \multicolumn{2}{c}{CL(min)} & \multicolumn{2}{c}{Accelerated EGO }\\
    \cmidrule(r){4-5}  \cmidrule(r){6-7}
    &  & & 10-point & 15-point & 10-point & 15-point  \\
    \hline
\multirow{2}*{Ackley10} & $M^{best}$ &  0.89 (0.8) & 1.23 (0.7) &
1.54 (0.8) &  0.89 (0.4) & 0.97 (0.2) \\
& time & 4793 (698) & 3574 (473) & 4057 (1254) & 412 (82) &  230 (22) \\
\hline
\multirow{2}*{Levy10} & $M^{best}$ & 2.86 (1.8) & 3.77 (1.1) & 6.80 (3.4) &  3.63 (1.3) & 5.54 (1.5) \\
& time & 5970 (776) & 5688 (1239) & 5099 (264) & 564 (168) & 365 (37)  \\ \hline
\multirow{2}*{Trid12} & $M^{best}$ & -347 (2) & -245 (54) & -230 (38)  & -251 (17) & -243 (51) \\
& time & 3042 (215) & 5512 (197) & 24116 (2105) & 399 (77)  & 306 (111) \\
	\hline
	\end{tabular}
    \begin{tablenotes}
    \linespread{1}
        \footnotesize
        \item[1] The total number of updating points is $150$ for every case.
        \item[2] The results are in the form of ``mean (standard deviation)".
        \item[3] The "min`` is the the minimum response value the algorithms can reach.
    \end{tablenotes}
\end{threeparttable}
\end{table}


Under high-dimensional case, accelerated EGO algorithm outperforms CL(min) significantly. From the minimum value the algorithms can obtain aspect, accelerated EGO can reach smaller response value than CL(min) algorithm for all three high-dimensional test functions with the same number of total runs and updating points per stage. For example, for Ackley10 function, the minimum value 10-point accelerated EGO obtains is $0.89$ on average, while the data of 10-point CL(min) is $1.23$. And the comparison is $-251$ to $-245$ for Trid12 function. From CPU time's point of view, much less CPU time is required for accelerated EGO compared to CL(min) algorithm. When parallel computation is available, reduction for the time of iterations is vital, especially under high-dimensional case. For example, with total $150$ updating points for Levy10 function, 10-point CL(min) algorithm costs $5688$ seconds (nearly $1.5$ hours) averagely to do the simulations, while accelerated EGO only requires $564$ seconds (nearly 10 minutes). Especially when the dimension and the number of updating points increase, this phenomenon becomes more highlighted. For example, for Trid function, the 15-point CL(min) algorithm run $24116$ seconds (nearly 7 hours) unexpectedly while accelerated EGO needs only $306$ seconds (nearly 5 minutes) on average. These results show that accelerated EGO is actually a computationally tractable method to parallelize EGO, especially under high-dimensional case.

Comparing with EGO algorithm, it validates again that accelerated EGO is not a strictly linear speedup algorithm. Under high-dimensional case, EGO can reach a smaller response value than accelerated EGO for most situations with the same number of total runs. For example, the minimum value of Levy10 function is $2.86$ averagely for EGO algorithm, while the 15-point accelerated EGO gets $5.54$. Only in one situation, accelerated  EGO obtains the same minimum value $0.98$ averagely with EGO algorithm when 10 updating points is selected per stage for Ackley10 function. Of course, when multiple points are updated per stage, the CPU time of accelerated EGO is much less than EGO with the same number of total runs. For example, for Trid12 function EGO algorithm runs $3042$ seconds to evaluate $150$ updating points, while 15-point accelerated iterates 10 times with $306$ seconds. But CL(min) does not have this advantage due to its hard computation for targeting the multi-ponts under high-dimensional case. Therefore, accelerated EGO would be a better choice if the the parallel computation can be conducted.

As discussed in previous sections, the searching speed of the accelerated EGO is much faster than other methods. Then we would like to know whether can the saved time be used to find better results.  That is, if the computation time of accelerated EGO is extended a little bit, whether can a much smaller response value be observed.  Actually, some additional simulations give the positive answer.  For example, for ``Ackley10" function, if we run the $10$-point accelerated EGO $5$ iterations more (i.e., $20$ iterations in total), the minimum response value is $0.63$ averagely, which is much smaller than the value $0.89$ of EGO algorithm. And the CPU time of accelerated EGO increases slightly being $448$ seconds averagely.  Similar results can be seen for other simulation cases.

\section{Application in hyper-parameter tuning for SVM}\label{sec:Hyper}

Recently, the task of choosing the best set of hyper-parameters for a machine learning model, i.e., a set of hyper-parameters that yields the best performance of the predictor on the available data set, is routinely facing with data scientists \citep{diaz2017}. For example, as for SVM with linear and nonlinear kernels \citep{cristianini2000}, one of the most promising learning algorithms for classification, selecting good hyper-parameters (including the regulation parameter $C$ and the Gaussian kernel parameter $\gamma$) to lead to a better generalization performances has become a major challenge at hand.

\begin{figure}[htbp]
	\centering
	\subfigure[german.n]{
		\includegraphics[width=3.7cm, height=3.6cm]{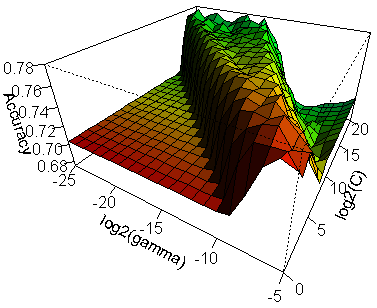}}
	\subfigure[svmguide2]{
		\includegraphics[width=3.7cm, height=3.6cm]{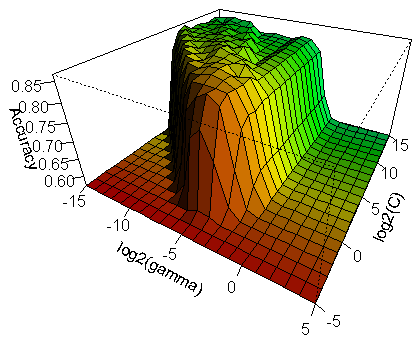}}
	\subfigure[glass]{
		\includegraphics[width=3.7cm, height=3.6cm]{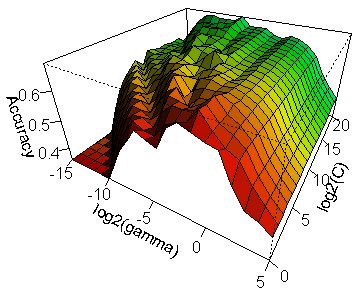}}
	\caption{Three examples of hyper-parameters search space}
	\label{fig:parameter_3D} 
\end{figure}

This problem can be treated as a black-box optimization situation for finding the global optimum of a function which is only vaguely specified and has many local optima. For example, as shown in \autoref{fig:parameter_3D}, we plots the $5$-fold cross validation average accuracy for three data sets, ``german.n", ``svmguide2" and ``glass", provided by \cite{chang2011} in a 3-dimensional surface, where $x$-axis and $y$-axis are $\log_2\gamma$ and $\log_2C$, respectively. The $z$-axis is the 5-fold cross-validation average accuracy. It is easy to see that there are many local optima and it is hard to obtain the global optimum.

Grid search is a widespread strategy for hyper-parameter optimization. It is obvious that the exhaustive grid search cannot effectively perform this task due to its high computational cost \citep{huang2007}. 
Here, we use the EGO, Constant Liar [CL(min)] and accelerated EGO algorithms to help the hyper-parameter turning for SVM and compare their performance with grid search.

\subsection{The procedure of hyper-parameter optimization}

\begin{figure*}[t]
	\centering
    \subfigure[Basic framework of SVM]{
        \includegraphics[width=10cm, height=4cm]{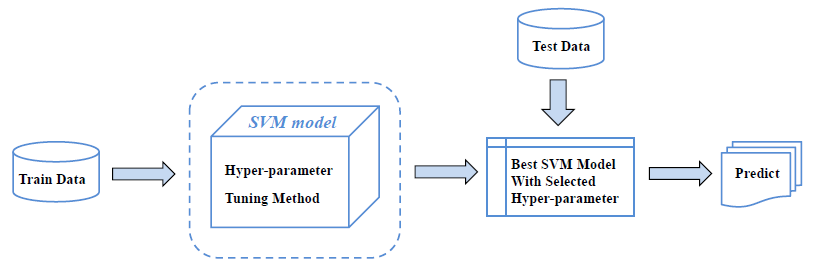}}
    \subfigure[The methods for hyper-parameter tuning]{
        \includegraphics[width=10cm, height=4cm]{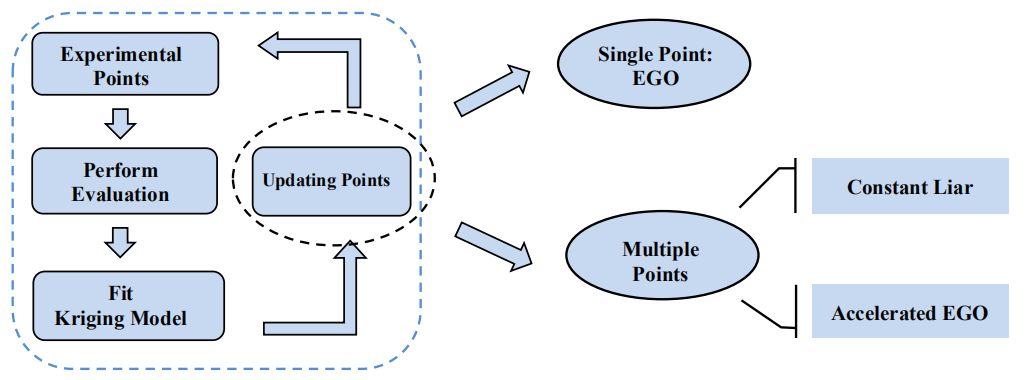}}
	\caption{The process of hyper-parameter optimization}
	\label{fig:HPO}
\end{figure*}

\autoref{fig:HPO} gives a visual representation of the process of hyper-parameter optimization for SVM.
The data set is divided into two part---Training data and Testing data---according to certain criterion. Then  the hyper-parameter optimization approaches are taken to select the best hyper-parameter combination for SVM based on the training data.

In the four methods for hyper-parameter tuning, a $k$-fold cross-validation is used to obtain the estimates of average accuracy (or generalization error) in the training data for each hyper-parameter combination. $k$-fold cross-validation is the most common performance assessment method to measure the performance for SVM. That is the training data is randomly split into $k$ mutually independent subsets with approximately equal sizes, then the $k-1$ subsets are employed to train the SVM model and the model is tested on the remaining one subset. This procedure is repeated $k$ times and the estimate of average accuracy is obtained by averaging the test accuracies over $k$ trials.

And for the surrogate-based methods,  EGO, CL(min) and accelerated EGO, the cross-validation average accuracy corresponding to each hyper-parameter combination is regarded as the response value to fit the Kriging model. Then the best hyper-parameter combination is obtained according to the three global optimization methods and is adopted to train SVM model with all the training data. And the resulting SVM model with the best hyper-parameters is used to predict the testing data and compute the prediction accuracy. Note that the testing data does not enter into the model training and hyper-parameter tuning procedure.

\subsection{Numerical results}

In this subsection, the EGO, CL(min) and our proposed accelerated EGO algorithms are applied to hyper-parameter optimization for SVM with different data sets: heart, german.n, australian, svmguide2 and glass.
These data sets are provided by \cite{chang2011}, and the details (classes, samples, features, etc.) of five data sets are shown in \autoref{tab:dataset}.
As the bench marker,  grid search method is also conducted.

\begin{table}[htbp]
	\centering
	\caption{Summary of data sets}
	\begin{tabular}{cccccc}
		\hline
		data set & class & sample & feature & \tabincell{c}{\thead{range of\\ $\log_2\gamma$}} & \tabincell{c}{\thead{range of \\ $\log_2C$}} \\
		\hline
		heart & 2 &  270 & 13 &  $[-20,\;0]$ & $[0,\;20]$  \\
	autralian & 2 &  690 & 14 &  $[-20,\;0]$ & $[-5,\;15]$ \\
 german.n & 2 & 1000 & 24 &  $[-25,-5]$  & $[0,\;20]$  \\
    svmguide2 & 3 &  391 & 20 &  $[-15,\;5]$ & $[-5,\;15]$ \\
		glass & 6 &  214 &  9 &  $[-15,\;5]$ & $[0,\;20]$ \\
	\hline
	\end{tabular}
	\label{tab:dataset}
\end{table}

Before hyper-parameter tuning, we randomly divide the data set into the training data set with $75\%$ samples and testing data set with the remaining samples and keep the four search methods using the same training and testing data.

\begin{sidewaystable*}
	\centering
	\caption{Result comparison of hyper-parameter optimization}
	\label{tab:svm_result}
\linespread{1.5}
\begin{threeparttable}
	\begin{tabular}{ccccccc}
		\hline
		\multirow{2}*{data set}&
		\multicolumn{3}{c}{Grid} &\multicolumn{3}{c}{EGO} \\
\cmidrule(r){2-4}  \cmidrule(r){5-7}
		&CV & predict & time & CV & predict & time \\
		\hline
		       heart & 0.853 (0.007) & 0.835 (0.015) & 52.82 (1.29) &  0.853 (0.594) & 0.843 (0.014) & 14.64 (2.22) \\
		  australian & 0.879 (0.002) & 0.849 (0.007) & 139.16 (0.96) & 0.877 (0.486) & 0.847 (0.007) & 15.29 (1.53)\\
		german.n & 0.770 (0.007) & 0.759 (0.009) & 1003.59 (34.84) & 0.769 (0.430) & 0.756 (0.006) & 30.12 (13.33)\\
		   svmguide2 & 0.850 (0.007) & 0.829 (0.013) & 82.28 (2.41) & 0.847 (0.705) & 0.831 (0.012) & 18.67 (3.09)\\
		       glass & 0.704 (0.017) & 0.733 (0.020) & 46.33 (2.12) & 0.699 (1.413) & 0.729 (0.029) & 14.26 (2.76)\\
		\hline
\multirow{2}*{data set}&
		\multicolumn{3}{c}{CL(min)} &\multicolumn{3}{c}{Accelerated EGO} \\
\cmidrule(r){2-4}  \cmidrule(r){5-7}
		&CV & predict & time & CV & predict & time \\
		\hline
		       heart & 0.851 (0.558) & 0.845 (0.019) & 11.71 (2.12) & 0.853 (0.553) & 0.841 (0.018) & 3.20 (0.56) \\
		  australian & 0.872 (0.433) & 0.841 (0.010) & 11.66 (1.68) & 0.872 (0.506) & 0.844 (0.008) & 2.76 (0.48) \\
		german.n & 0.769 (0.382) & 0.757 (0.007) & 11.21 (2.27) & 0.771 (0.422) & 0.757 (0.008) & 2.64 (0.58)\\
		   svmguide2 & 0.842 (0.721) & 0.828 (0.013) & 17.04 (2.71) & 0.844 (0.670) & 0.826 (0.014) & 3.66 (0.69)\\
		       glass & 0.695 (1.393) & 0.733 (0.029) & 12.86 (1.54) & 0.698 (1.268) & 0.736 (0.028) & 3.47 (0.76)\\
		\hline
	\end{tabular}
    \begin{tablenotes}
        \footnotesize
        \item[1] The results are in the form of ``mean (standard deviation)".
        \item[2] The ``CV" means the cross-validation average accuracy.
        \item[3] The ``predict" is the predicting accuracy of the testing data set.
    \end{tablenotes}
\end{threeparttable}
\end{sidewaystable*}

For grid search method, $441$ hyper-parameter combinations are searched and the hyper-parameter pair with the largest $5$-fold cross-validation average accuracy among them is chosen. Then this selected hyper-parameter pair is to train SVM with all the training data and the predicting accuracy of the testing data can be obtained.

For EGO, CL(min) and accelerated EGO algorithms, a uniform design with $21$ points is set to be the initial design and $5$ updating points are selected at each iteration for CL(min)and accelerated EGO. We run EGO algorithm $20$ iterations, Constant Liar and accelerated EGO $4$ iterations, i.e., all algorithms are tuning $41$ hyperparameters including the initial experimental points. For accelerated EGO, the initial QMC point pool is the sobol sequence with size $100$. In addition, $5$-fold cross validation is used to evaluate the performance of SVM model.

In the process of hyper-parameter optimization, in order to realize the `real' parallelization of Constant Liar and accelerated EGO algorithms, the R packages `` foreach" \citep{foreach2019} and ``doParallel" \citep{doPar2019} are employed to do the parallel computing. Specifically speaking, at each iteration, the evaluations of the $5$ updating experimental points are simultaneously done on $5$ different cores of the computer respectively, i.e, the $5$-fold cross-validation average accuracy corresponding to the $5$ updating hyper-parameter combinations are obtained simultaneously.

\autoref{tab:svm_result} demonstrates the comparison among the results of grid search, EGO, CL(min) and the proposed accelerated EGO methods for each data set using the SVM classifier. In each methods, the largest cross validation accuracy denoted by ``CV" and the prediction accuracy of the testing data set denoted by ``predict" are listed. In addition, the CPU time (in seconds) charged for the execution of user instructions of the hyper-parameter optimization process are also recorded. For grid search $10$ replications are run due to its high computation cost, and others run $100$ replications. The results are shown in the form of ``mean (standard deviation)".

The grid search can get the best cross validation accuracy for most data sets due to its exhausted computation, however other methods can be little short of grid search for their less computation and the nearly same accuracy. Comparing among EGO, CL(min) and accelerated EGO, EGO can get the best cross validation accuracy for all the data set except ``german.n". For this ``german.n" data set, the cross validation accuracy of accelerated EGO is $0.771$ averagely which is the largest among these three algorithms. And accelerated EGO performs better than CL(min) from the cross validation accuracy aspect for most data set, for example, for ``heart", the cross validation accuracy of accelerated EGO is $0.002 (=0.853-0.851)$ more than CL(min) averagely.

It can be observed that, thanks to the parallel computation provided by R packages, the CPU time for the hyper-parameter optimization process of CL(min) and accelerated EGO are less than EGO algorithm. And for accelerated EGO, much significant improvement can be observed. For example, for ``australian" data set, the CPU time of EGO is  $5.5\;(\approx15.29/2.76)$ times as much as accelerated EGO, and the CL(min) gets little reduction of the CPU time with $11.66$ seconds since it may cost a lot to find the multiple updating points per cycle. That is to say, accelerated EGO indeed speeds up and lessens the computation burden greatly for targeting the multiple updating points per iteration.

In addition, if we conduct additional experiments for this situation, much better results can be recorded. If $8$ more iterations is added for accelerated EGO algorithm, i.e., we run this algorithm $12$ iterations ($60$ updating points totally), for ``australian" , the CPU time of the hyper-parameter tuning process is $9.48$ seconds on average which is still less than that ($15.29$ seconds) of EGO. And the cross validation accuracy of accelerated EGO is $0.879$ averagely, which is the same as the result of exhausted grid search and is larger than that ($0.877$) of EGO. Similar results can be observed for ``svmguide2" and `` glass" data set.

\section{Conclusion and discussion}\label{sec:con}

In this paper, a novel batch sequential adaptive design method, accelerated EGO, is proposed to parallelize EGO algorithm and to release the computation burden for targeting the multiple updating points per stage. This algorithm employs the techniques of randomized quasi-Monte Carlo and resampling to select the updating points and it is a fast way to obtain samples with good representativeness and large EI values.

The efficiency of the proposed  accelerated EGO algorithm is validated over nine test functions with dimension from 2 to 12. Summarized the empirical computations, we can get following conclusions:
\begin{enumerate}
	\item Accelerated EGO is a batch version of ordinary EGO algorithm. The results show accelerated EGO indeed speeds up the convergence of EGO algorithm and saves the cost of stages. Further more, the empirical simulation also show that the saved time can be changed to the better results.
	
    \item Compared with the counterpart parallel EGO---Constant Liar algorithm, much significant improvement can be observed. Under low-dimensional case, accelerated EGO performs similarly with Constant Liar with less computational burden to select the multiple updating points. However, accelerated EGO can get more acceptable optimum value than Constant Liar algorithm under high-dimensional case. And in this situation, the reduction of computation cost for accelerated EGO is much highlighted.
\end{enumerate}

In addition, EGO, Constant Liar and accelerated EGO algorithms are applied to hyper-parameter optimization for SVM. $5$ data sets are used to compare these algorithms. The results show with the real utility of parallel computational technique, accelerated EGO is more efficient than EGO and releases the computational burden to target the multiple updating points per iteration dramatically.




\section*{Funding}
This study is partially supported by the National Natural Science Foundation of China (No.11571133 and 11871237).

\bibliographystyle{tfs}
\bibliography{tex}

\end{document}